\documentclass[sigconf]{acmart}

\AtBeginDocument{%
  \providecommand\BibTeX{{%
    \normalfont B\kern-0.5em{\scshape i\kern-0.25em b}\kern-0.8em\TeX}}}

\usepackage{xcolor}
\usepackage{amsthm}

\theoremstyle{definition}
\newtheorem{exmp}{Example}[section]
\usepackage{amsmath}
\usepackage{algorithm}
\usepackage[noend]{algpseudocode}

\usepackage{hyperref}

\makeatletter
\def\BState{\State\hskip-\ALG@thistlm}
\makeatother

\setcopyright{acmcopyright}
\copyrightyear{2021}
\acmYear{2021}
\setcopyright{acmlicensed}\acmConference[KDD '21]{Proceedings of the 27th ACM SIGKDD Conference on Knowledge Discovery and Data Mining}{August 14--18, 2021}{Virtual Event, Singapore}
\acmBooktitle{Proceedings of the 27th ACM SIGKDD Conference on Knowledge Discovery and Data Mining (KDD '21), August 14--18, 2021, Virtual Event, Singapore}
\acmPrice{15.00}
\acmDOI{10.1145/3447548.3467088}
\acmISBN{978-1-4503-8332-5/21/08}

\begin{document}
% \fancyhead{}
\title{AutoSmart: An Efficient and Automatic Machine Learning framework for Temporal Relational Data}


\author{Zhipeng Luo\textsuperscript{1}, Zhixing He\textsuperscript{1}, Jin Wang\textsuperscript{1}, Manqing Dong\textsuperscript{1}$^\dag$, Jianqiang Huang\textsuperscript{2}, Mingjian Chen\textsuperscript{2}, Bohang Zheng\textsuperscript{2}
}
\authornote{$^\dag$ Manqing Dong is the corresponding author.}
\affiliation{
\institution{\textsuperscript{1}DeepBlue Technology, Beijing, China\\ \textsuperscript{2}Peking University, Beijing, China}
\country{}
}
\email{{luozp,hezhx,wangjin,dongmq}@deepblueai.com, {jqhuang,milk,wenwuhang}@pku.edu.cn}

\renewcommand{\shortauthors}{Z. Luo et al.}

\begin{abstract}
Temporal relational data, perhaps the most commonly used data type in industrial machine learning applications, needs labor-intensive feature engineering and data analyzing for giving precise model predictions. An automatic machine learning framework is needed to ease the manual efforts in fine-tuning the models so that the experts can focus more on other problems that really need humans' engagement such as problem definition, deployment, and business services. However, there are three main challenges for building automatic solutions for temporal relational data: 1) how to effectively and automatically mining useful information from the multiple tables and the relations from them? 2) how to be self-adjustable to control the time and memory consumption within a certain budget? and 3) how to give generic solutions to a wide range of tasks? In this work, we propose our solution that successfully addresses the above issues in an end-to-end automatic way. The proposed framework, AutoSmart, is the winning solution to the KDD Cup 2019 of the AutoML Track, which is one of the largest AutoML competition to date (860 teams with around 4,955 submissions). The framework includes automatic data processing, table merging, feature engineering, and model tuning, with a time\&memory controller for efficiently and automatically formulating the models. The proposed framework outperforms the baseline solution significantly on several datasets in various domains. The source code is available at \url{https://github.com/DeepBlueAI/AutoSmart}.
\end{abstract}



\maketitle

%%%%%%%%%%%%%%%%%%%%%%%%%%%%%%%%%%%%%%%%%%%%%%%%%%%%%%%%%%%%%%%%%%%%%%%%%%%%%%%%%%%%%%%%%
\section{Introduction}
% Machine learning
As one of the most commonly used data types in the database, temporal relational data is widely distributed in real-world applications, such as online advertising, recommender systems, financial market analysis, medical treatment, fraud detection, etc~\cite{zhou2020kdd}.
Along with multi-perspective relational information through nested tables, chronological information contains rich information that can be exploited to improve machine learning performance.
Currently, the analysis of temporal relational data is often carried out by experienced human experts with in-depth domain knowledge in a labor-intensive trail-and-error manner~\cite{Hutter2014An}. 
Therefore, a reliable automatic solution on temporal relational data is urgently needed.
In recent years, academia and the industrial community have made promising progress on a few components of Automatic Machine Learning (AutoML)~\cite{zoller2019benchmark}, such as automatic algorithm selection and hyper-parameter optimization~\cite{baker2016designing}. Nonetheless, the challenge of building an end-to-end and fully unmanned pipeline is still far from being solved~\cite{hassan2020level}. In addition, this progress is especially beneficial to neural architecture search~\cite{baker2016designing} for automatic learning on deep methods; while temporal relational data, as a sub-type of tabular data, has less benefit from the trend of deep learning and its automation. 

\begin{figure}[t]
    \centering
    \includegraphics[width=\linewidth]{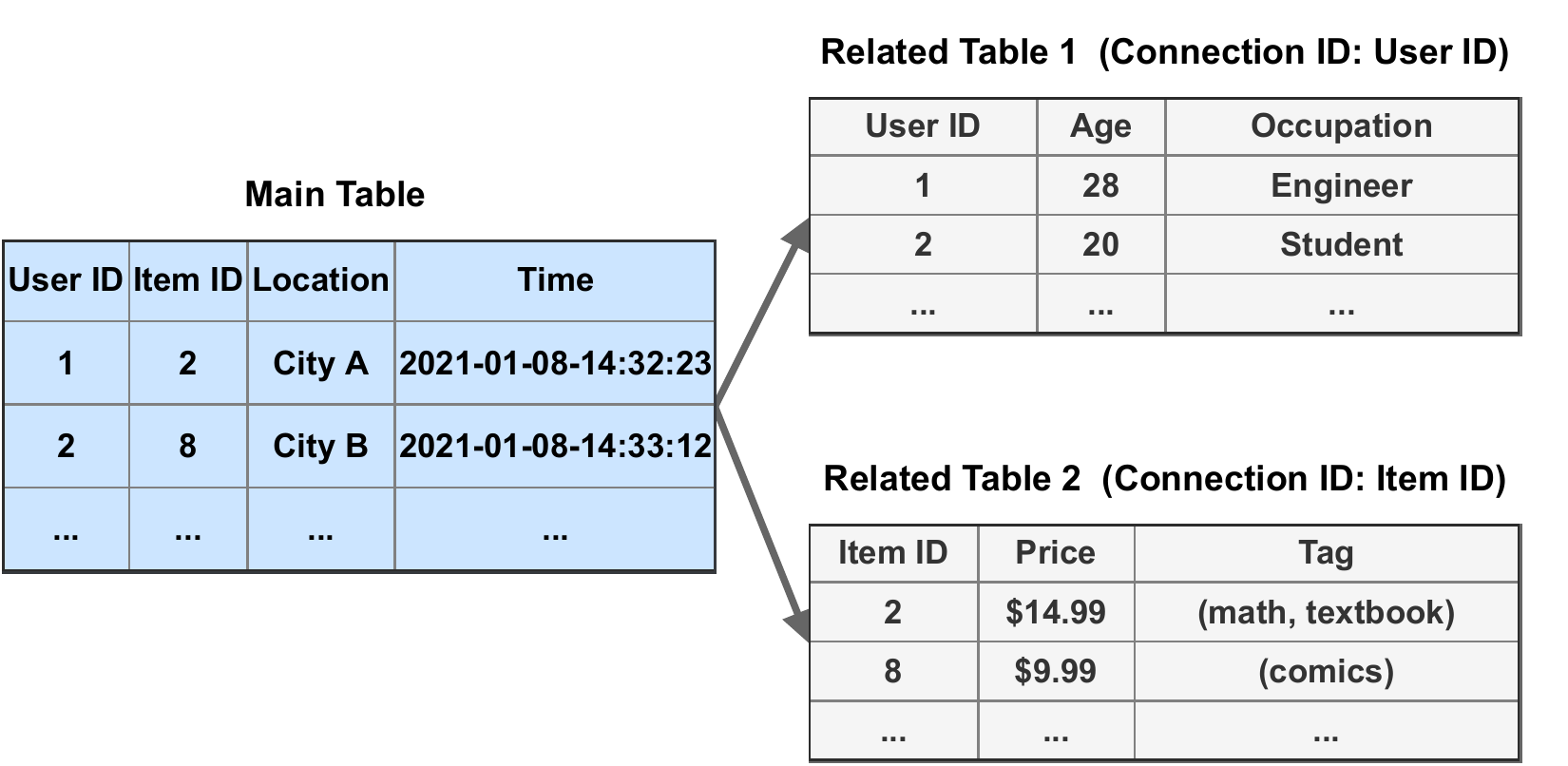}
    \caption{An example of temporal relational data in a real world context.}
    \label{fig:example_multi-table}
\end{figure}

There are several challenges in building an automated machine learning framework for temporal relational data. 
\textbf{First}, conventional ways in analyzing temporal relational data requires task-specific domain knowledge for feature engineering; therefore, implementing a generic automated feature engineering strategy to extract useful patterns from the temporal relational data is challenging. An example for temporal relational data in a real-world context is illustrated in Figure~\ref{fig:example_multi-table}. The challenges lie in: efficiently merging the information provided by multiple related tables into a concentrated format, automatically capturing meaningful interactions among the tables, effectively preserving the valuable, temporal, and cross-table information, and automatically avoiding data leak when the data is temporal. 
% Second
\textbf{Second}, computational resources to implement AutoML approaches still have expected limits in runtime and memory consumption even in the presence of ever-increasing computing power. For the real-world task with complicated and scattered data sources, the performance of ML applications relies heavily on how to utilize the temporal relational data productively. Accurately controlling runtime and computational resources plays a vital role in evaluating the business value of the ML model. 
For example, to plug data from multiple data partnerships into a commercial analytical platform, data providers and vendors may come in hundreds and change monthly. There is a need for quick and efficient integration in such cases, where some parts can be automated. 
\textbf{Third}, Providing generic automated solutions for temporal relational data is challenging, especially when adapting automated models to unseen tasks in different domains.
To the best of our knowledge, existing AutoML frameworks~\cite{zoller2019benchmark} for temporal relational data still require human assistance for providing a feasible solution. Therefore, an end-to-end AutoML solution that can automatically adapt to various situations is very much needed.
% To provide a AutoML which can automatically adapt to various situations  end to end solution is very much needed. 

For solving the above challenges, we introduce our winning solution, AutoSmart, an efficient and automatic machine learning framework that supports temporal relational data. 
The proposed framework is composed of four automatic modules, including data preprocessing, table merging, feature engineering, and model tuning, along with a time\&memory controller for controlling the time and memory usage of the model.
The experiments are implemented on the competition data in KDD Cup 2019, which includes five public-available datasets and five unseen datasets in different application domains. The proposed framework shows consistent outstanding performance on all ten datasets while efficiently managing both time and memory consumption. 

Specifically, we make the following contributions:
\begin{itemize}
    \item The proposed framework performs well on time relational data in an end-to-end automatic way. Our automatic feature engineering strategy uses four sequential feature generation and feature selection modules, which can effectively learn valuable information from the input data regardless of the data domains.
    \item Data sampling, code optimization, and risk assessment are implemented for controlling and monitoring the time and memory usage of the framework thoroughly, leading to an excellent performance on the given data with fast adapt to the time and memory budget.
    \item Experiments on ten datasets from different domains suggest the effectiveness and the efficiency of the proposed framework. We further conduct experiments dealing with rare conditions, such as larger datasets, different types of missing values, and strictly restricted time limits, where the proposed framework can be smoothly adapted to other scenarios and retain stable performance.
    % Our further experiments on system testing demonstrate that the framework also works well on rare conditions, such as dealing with larger datasets, different types of missing values, and strictly restricted time limits, indicating our framework can be smoothly adapted to other scenarios. 
    We make the code publicly available for easier reproduction: \url{https://github.com/DeepBlueAI/AutoSmart}.
\end{itemize}

%%%%%%%%%%%%%%%%%%%%%%%%%%%%%%%%%%%%%%%%%%%%%%%%%%%%%%%%%%%%%%%%%%%%%%%%%%%%%%%%%%%%%%%%%
\section{Related Work}

\begin{figure*}[htbp]
\includegraphics[width=\textwidth]{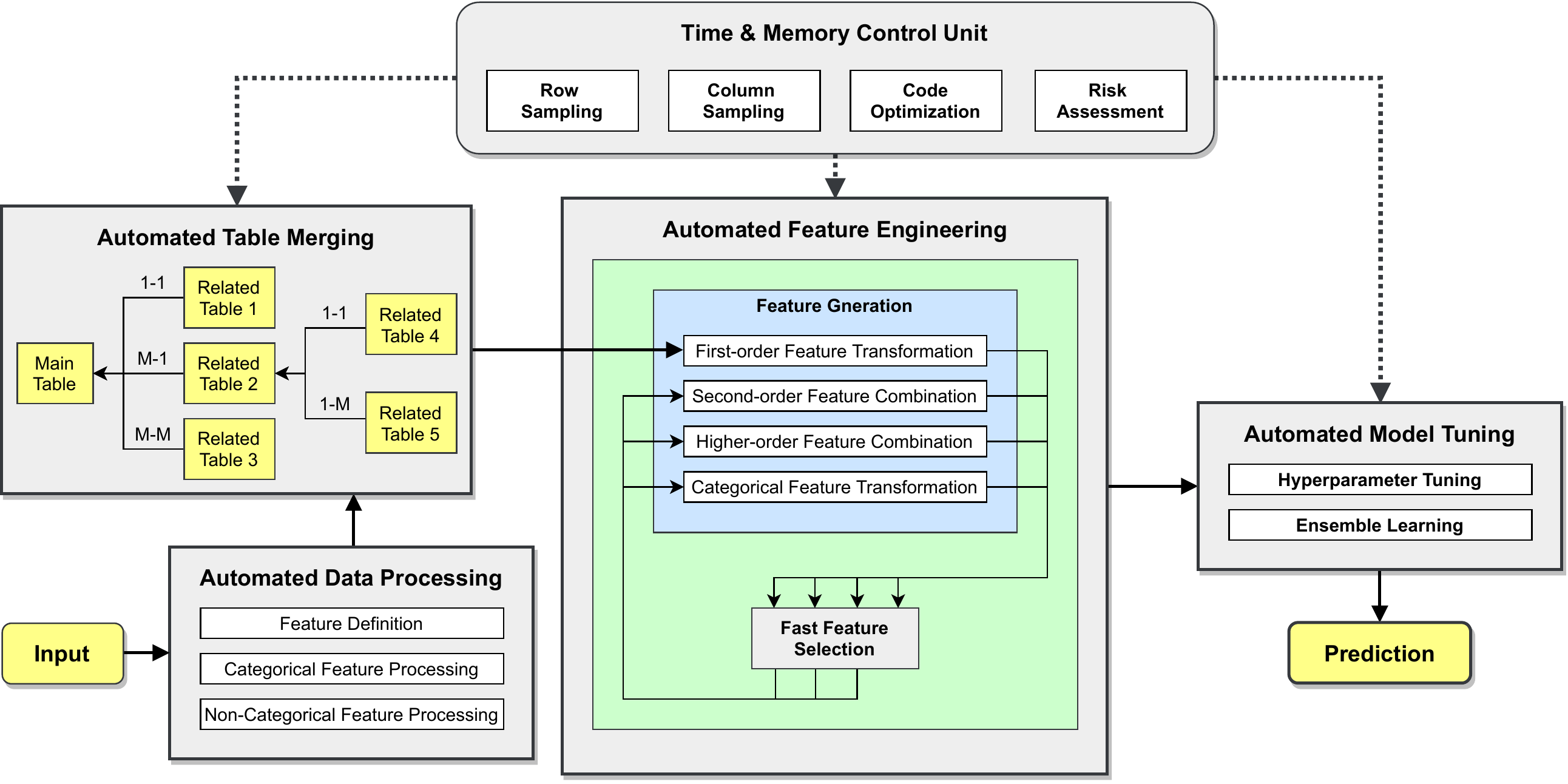}
\caption{A schematic illustration of the architecture of the AutoSmart, which composes of automated data preprocessing, automated table merging, automated feature engineering, and automated model tuning, with a time \& memory control unit to control the resource usage.}
\label{fig:model_framework}
\end{figure*}

Researches for automated machine learning (AutoML) has increasingly emerged in the last decade to ease the manual efforts in most aspects of machine learning, such as feature generation, feature selection, and model learning~\cite{Drori2019AutoML,Wang2019ATMSeer}. 
A general target for AutoML is automatically learning a model to best fit the data in an end-to-end manner, which includes three sub-goals: good model performance, less assistance from human experts, and high computational efficiency. 
Two critical components in formulating an AutoML framework are feature engineering and model learning; a general strategy is to use the features generated by the former as the inputs for the latter. 
Many existing works focus on providing robust solutions to either automatic feature engineering or automatic model learning. 

\paragraph{Automatic feature engineering.}
It is widely believed that data and features determine the upper bound of machine learning, and models and algorithms can only approach this limit~\cite{graepel2012ml} -- this suggests that the performance of a machine learning algorithm heavily depends on the quality of input features and the effectiveness of feature engineering.
While recent developments in automatic processing of images~\cite{wong2018transfer}, texts~\cite{estevez2019automl}, and signals~\cite{he2018amc} by deep learning methods, feature engineering for relational data remains iterative, human-intuition driven, and hence, time-consuming. 
% Besides, deep learning-based AutoML methods lie on parameter searching space rather than automatic feature engineering, where the parameter scale of a neural network is normally larger than the other light-weighted methods, which make deep learning based methods take more time and memory cost on relational data. 
Generally, feature engineering for relational data includes feature generation and feature selection. The former creates new features through a functional mapping of the original features or by discovering hidden relationships between the original features~\cite{Shaul2002Feature}. The latter tries to remove the redundant, irrelevant, or misleading features generated by the former~\cite{Shaul2002Feature,koppen2000curse}, which tends to simplify the model and avoid over-fitting caused by the 'curse of dimensionality'~\cite{He2019AutoML}.
The two processes are typically iteratively combined to find the optimum feature subset efficiently. 
If we consider the generation process of one feature as the transformation through an operator, then the whole process of generating all features can be treated as a node selection problem in a transformation tree: the root node represents the original features; each edge applies one specific operator leading to a transformed feature set~\cite{Khurana2016Cognito,lam2017one}. Simple exhaustive approaches evaluate the performance through all nodes or randomly selected nodes from a fully expanded transformation tree~\cite{dash1997feature}. However, the process for such feature engineering approaches may be highly resource-consuming because of the extensive and complex $O(2^N)$ search space.
Instead of iteratively exploring the transformation tree, improved exhaustive approaches stopped at a predefined depth~\cite{Kanter2015Deep, Katz2016ExploreKit}, in line with the idea that higher-order features are less likely to contain useful information. Greedy search~\cite{Khurana2016Cognito, Margaritis2009Toward} is another approach to reduce the exponential time complexity but retaining an acceptable performance. For instance, Margaritis~\cite{Margaritis2009Toward} used a combination of forward and backward selection to select a feature subset. Recently, dedicated feature selection approaches are used in the middle of the feature generation process in helping the search for optimal features. Cognito~\cite{Khurana2016Cognito} handles the scalability problem of feature selection on large datasets successfully by applying feature filtering on every feature node during traversal.  
As for relational data, Deep Feature Synthesis algorithm~\cite{Kanter2015Deep} spend more effort on extracting the valuable information based on the interactions between tables in a cascade manner, i.e. order-wise generating relational features from simple to complex.

\paragraph{Automatic model learning.} Once chosen a model, hyper-parameter tuning strategies, such as grid search, random search, may significantly boost the performance~\cite{yu2020hyper}. 
However, the ample search space (especially for the complex model with many parameters) prevents these methods from simply applying to the AutoML with a limited resource budget.  Therefore, an efficient hyperparameter searching method is required. 
Bayesian optimization~\cite{snoek2012practical, yu2020hyper} is widely used in a sequential model-based algorithm configuration (SMAC)~\cite{shahriari2015taking} to achieve a faster converging rate, from using standard Gaussian Process as the surrogate model to random forests or the Tree Parzen Estimator~\cite{bergstra2011algorithms}.
The result of the AutoML model can be further improved by combining the prediction from multiple models. Choosing the best model based on the metrics on the validation set or ensembling different types of models with multi-layer stacking model~\cite{ledell2020h2o, erickson2020autogluon, sagi2018ensemble} are commonly used to raise the performance and the robustness of the final result.

\paragraph{AutoML frameworks.} Many existing works have made great efforts on providing an end-to-end solution for AutoML from both academic and industrial communities. 
Some famous frameworks tackling tabular data include: 
Google AutoML tables~\cite{bisong2019overview}, Azure Machine Learning~\cite{klein2017azure}, Auto-Sklearn~\cite{feurer2019auto}, TPOT~\cite{olson2016tpot}, and AutoGluon-tabular~\cite{erickson2020autogluon}. Although these frameworks achieve competitive results~\cite{Marc2020Benchmark} on selected datasets, few of them support temporal relational data. In addition, they pay more attention to model selection than feature engineering; therefore, the model becomes inefficient in handling the datasets containing a large amount of concealed information.
% Here we propose an efficient and effective framework to address the above issues for the temporal relational data, with time and memory control in the whole training process.

%%%%%%%%%%%%%%%%%%%%%%%%%%%%%%%%%%%%%%%%%%%%%%%%%%%%%%%%%%%%%%%%%%%%%%%%%%%%%%%%%%%%%%%%%
\section{Method}
In this work, we propose a general AutoML framework that supports temporal relational data, including automatic data preprocessing, automatic table merging, automatic feature engineering, and automatic model tuning, along with a time \& memory control unit. A systematic illustration of the proposed framework is shown in Figure~\ref{fig:model_framework}.

\subsection{Problem definition}
Temporal relational data usually stands for multiple relational tables, i.e., a main table describing temporal information about the key IDs with several related tables containing auxiliary information about the key IDs. The \textit{key IDs} are the connection columns between the main table and the related tables. 
An example of the temporal relational tables is shown in Figure~\ref{fig:example_multi-table}, where the \textit{key IDs} here are \textit{User ID} and \textit{Item ID}.
Suppose the \textit{main table} is denoted as $T_{0}$, the $K$ \textit{related tables} are denoted as $\{T_{k}|k\in (1, ..., K)\}$, then the problem can be defined as: given a main table $T_{0}$, the labels $Y$ for the main table, and its related tables $\{T_{k}|k\in (1, ..., K)\}$, the goal is predicting the labels for a new table in the same format as $T_{0}$. 
Each table contains several columns describing either numerical features, categorical features, multi-categorical features (e.g. the cast list for a movie), or temporal features (e.g. a specific time). 

\subsection{Automated Data Preprocessing}
Figure~\ref{fig:example_multi-table} gives an example of the temporal relational data in a real-world context for easier understanding of our target. 
However, due to the privacy policy, the provided data may be encrypted, and the real meanings of the features are unavailable (refer to Figure~\ref{fig:example_x-1} and ~\ref{fig:example_x-M}, the data provided in the 2019 KDD Cup AutoML track).
To provide a general AutoML solution to fit more scenarios, we focus on mining the information from the feature types and the relations between the features.

Among the above mentioned four basic types of features, we define three new types, namely 'factor', 'key', and 'session' features, as \textit{base features}, which we found containing essential information about the multiple tables and is most helpful for the final prediction.
Recall that each dataset has \textit{key IDs} linking the main table with the related tables, we define those \textit{key IDs} as '\textbf{key}' features for each table. '\textbf{Factor}' feature is selected from the 'key' features, where the 'key' feature with more unique values is identified as the 'factor' feature. The '\textbf{session}' features are selected from the categorical features with rules that we hope each '\textbf{session}' feature can efficiently depict each 'factor', e.g. a 'factor' may contain several values in the 'session' feature while those values are unique for this 'factor' and will not be used for describing other 'factor's. 

\begin{exmp}
Suppose we are under a recommendation scenario, the '\textbf{factor}' feature can be regarded as user ids. The main table may contain the basic information about the user and the item information such as the time of the user staying on the page and the price of the item; the label $y$ here could be the user's click behavior (click -> 1 or not -> 0) on the item. The '\textbf{session}' feature behaves like the identical information of a particular period, such as the user IP addresses, where the user may buy many items at the same IP address. The '\textbf{key}' features, in this case, may include the user ids and the item ids. Then the related tables may contain additional information describing the users or the items, such as the user profile and the item descriptions. 
\end{exmp}

Besides defining those three features, we remove less informative features, e.g. the numerical features with very small variance, and the categorical features with single values or almost all different values.
Then, we preprocess the features according to the feature type.

\paragraph{Preprocessing for categorical and multi-categorical features.} 
Two common ways for preprocessing the categorical features are encoding the values (e.g. string) to integers and one-hot vectors. 
The latter strategy is not considered in our system due to the high computation cost.
A traditional way for encoding the categorical values to integers is using a set of sequential integers like (0, 1, 2) to represent the set ('apple', 'banana', 'peach'). However, in case a value (e.g. 'apple') appears in both a categorical feature (a single 'apple' value) and a multi-categorical feature (e.g. the set ('apple', 'banana')), this encoding strategy may use different integers to represent the same value. 
For solving this issue, we design a recursion function to automatically allocate the related categorical and multi-categorical features into the same blocks.
To be more specific, we initialize a feature pair diagonal matrix, denoted by $M$, with each row and column represents a categorical feature or multi-categorical feature, where $\{M_{i,j}=1|i=j\}$ and the others are assigned to 0. Then, we calculate the intersection of two sets of values in categorical or multi-categorical features and set $M_{i,j}=1$ if feature $i$ and feature $j$ have an overlap larger than a designed threshold (e.g. 10\%). 
Algorithm~\ref{alg:mc_graph} gives the pseudo-code of how we separate the blocks recursively. Then, the encoding will be conducted within each block to preserve cross-feature information.

\begin{algorithm}
\caption{Automated Feature Blocks Generating}\label{alg:mc_graph}
\begin{algorithmic}[1]
% \Procedure{MyProcedure}{}
\Require{N categorical features}
\Ensure{A block dictionary $B$ with $b$ features blocks}
\State Initialize a zero matrix $M\in \mathbb{R}^{N\times N}$ with diagonal values as 1
\For{$i \in range(N)$}
\For{$j \in range(N)$}
\State Set $M_{i,j}=1$ if feature $i$ and $j$ have much overlaps
\EndFor
\EndFor
\State Initialize a candidate set $c\in \mathbb{R}^{N}$ with zero values
\State Initialize a block dictionary, set the initial block id as $b=0$
\For{Feature $n \in range(N)$}
\If{Feature $c_{n}=1$}
\State Continue
\EndIf
\State Update candidate set with $c_{n}=1$
\State Update block id $b = b+1 $
\State Initialize block set $B_{b}=\{\}$
\State \Call{SearchID}{Feature $n$, Block set $B_{b}$, Candidate set $c$}
\EndFor
\Function{SearchID}{Feature $i$, Block set, Candidate set}
\State Append feature $i$ to the current block set
\For{$j \in range(N)$}
\If{$c_{j}$ = 0}
\If{$M_{i,j}=1$}
\State Update candidate set with $c_{j}=1$
\State \Call{SearchID}{Feature $j$, Block set, Candidate set}
\EndIf
\EndIf
\EndFor
\EndFunction
\end{algorithmic}
\end{algorithm}

\paragraph{Preprocessing for numerical features and temporal features.} We keep the raw values for those features in the tables and sort the table according to the temporal features. For saving the computation cost, we use smaller data type to represent the numerical values (e.g. \textit{float32} to \textit{float16}) if possible.

\subsection{Automated Table Merging}
A typical structure for temporal relational data is shown in Figure~\ref{fig:example_multi-table}, where there is a main table storing basic information about the \textit{key IDs} and several related tables storing auxiliary information.
The tables are linked with the \textit{key ids} in either one of the four following situations:
\begin{itemize}
    \item One-to-One (1-1): one row in table A may be linked with only one row in table B and vice versa.
    \item Many-to-One (M-1): one row in table A is linked to only one row in table B, but one row in table B may be linked to many rows in table A.
    \item One-to-Many (1-M): one row in table A may be linked with many rows in table B, but one row in table B is linked to only one row in table A.
    \item Many-to-Many (M-M): one row in table A may be linked with many rows in table B and vice versa.
\end{itemize}
Our target is merging all additional information in formulating a new informative main table for prediction. 

\begin{figure}
\center
\includegraphics[width=0.95\linewidth]{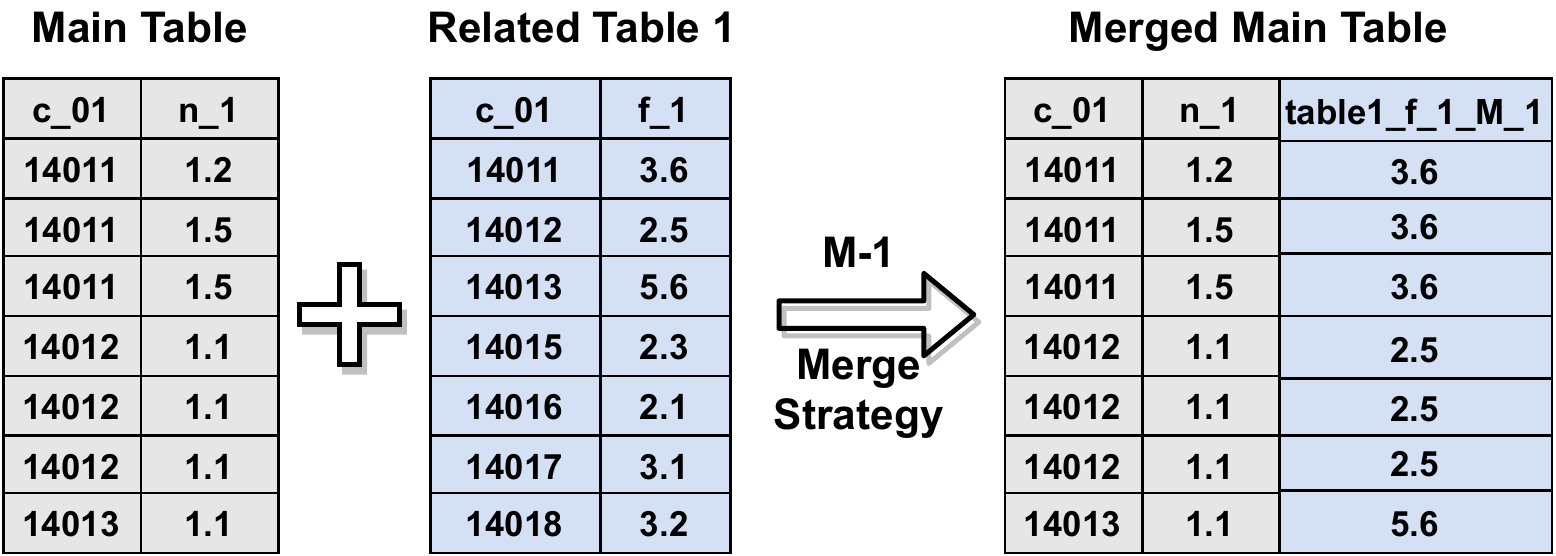}
\caption{An illustration of table merging under the relationship of M-1. To merge the two tables, we directly join the column $f\_1$ from the related table to the main table according to the \textit{key id} $c\_01$ in the two tables.}
\label{fig:example_x-1}
\end{figure}

For the cases the connection types are 1-1 and M-1, it is relatively easy to merge the related table with the main table: concatenate the single features in the related tables to the according \textit{key IDs} in the main table. 
An example for the M-1 relationship is shown in Figure~\ref{fig:example_x-1}. The connection \textit{key id} is $c\_01$. Taking the key value $14011$ as an example, the value of feature $f\_1$ in the related table is 3.6. Then after table merging, the value for key value $14011$ in new column $f\_1$ is 3.6.

\begin{figure}
\center
\includegraphics[width=0.95\linewidth]{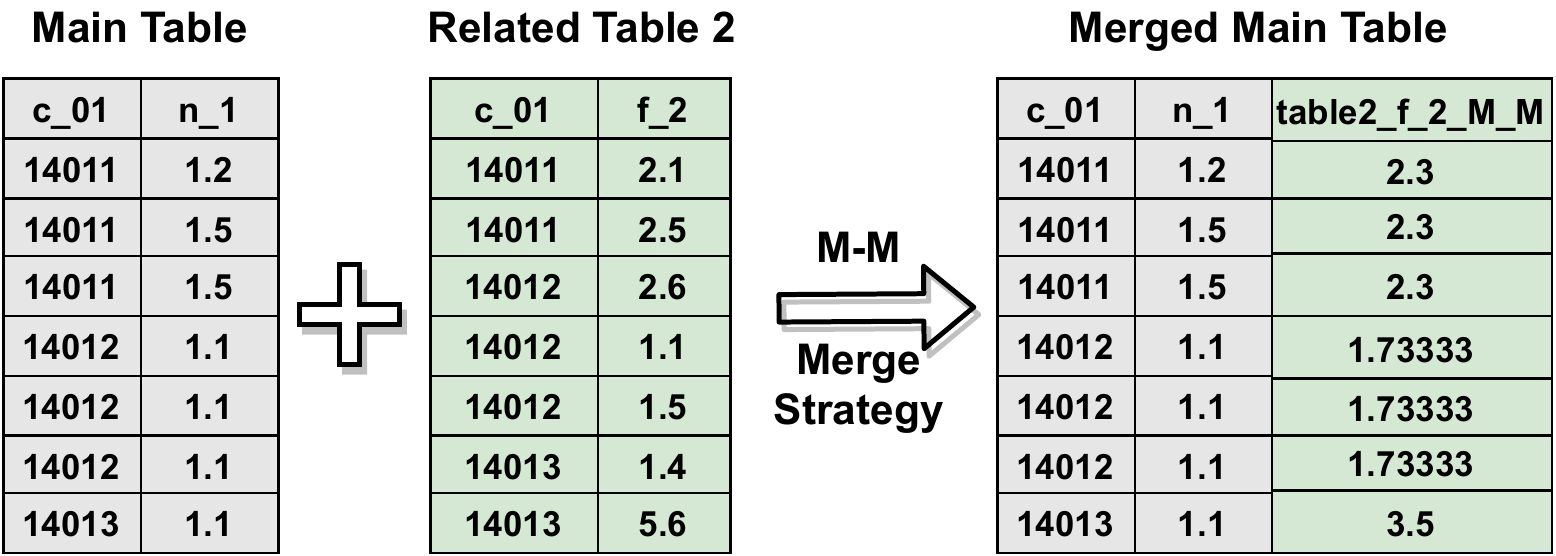}
\caption{An illustration of table merging under the relationship of M-M. To merge the two tables, we take the mean values (for numerical features) in related tables as the merged values to the main table, according to the \textit{key id} $c\_01$.}
\label{fig:example_x-M}
\end{figure}

For the cases the connection types are 1-M and M-M, we merge the related table according to the feature type, i.e. categorical or multi-categorical features, numerical features, or temporal features. For example, for numerical features and categorical features, we take mean values and mode values in the related table as the value for the \textit{key ids} in the main table. 
As for the temporal features, we take the newest time as the merging value for the main table. 
An example is shown in Figure~\ref{fig:example_x-M}, where $c\_01$ is the \textit{key id} and $f\_2$ is the merging feature. Taking the key-value $14011$ as an example, the value for this key in the merging table is the mean value, i.e. 2.3, of the feature $f\_2$ in the related table.

\subsection{Automated Feature Engineering}
\begin{figure}
\includegraphics[width=0.8\linewidth]{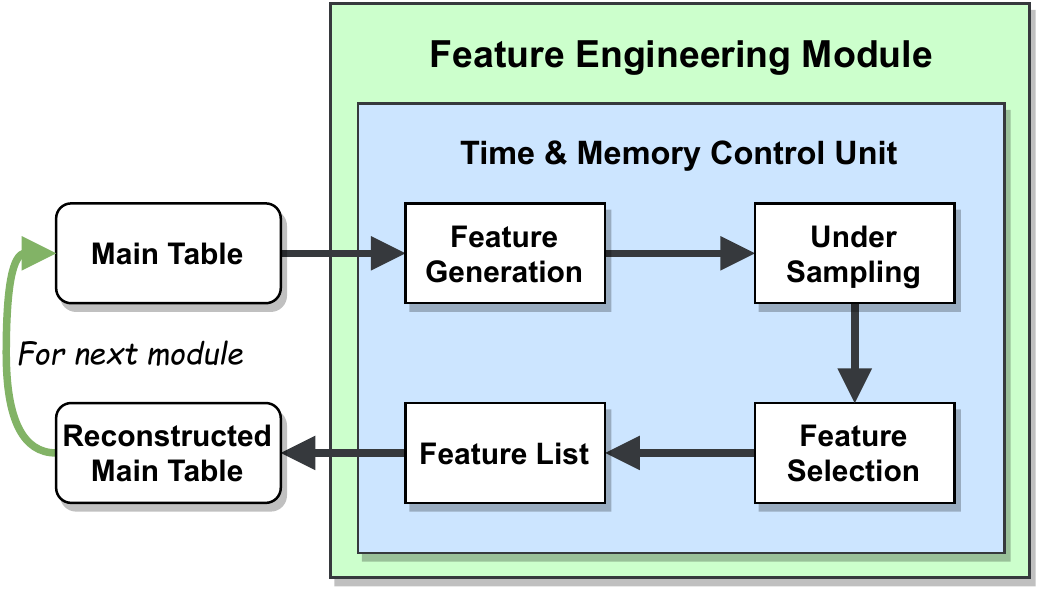}
\caption{An illustration of the iterative processes in the automatic feature engineering module.}
\label{fig:feature_engineering}
\end{figure}

We divide feature engineering into four sequential modules to take full usage of table information and minimize the memory usage. For each module, we use LightGBM~\cite{ke2017lightgbm} to verify the effectiveness of the features and do feature selection.
We will first introduce the details about the four feature modules and then our strategies for feature selection.

\subsubsection{Feature Generation}
In particular, we consider the following four groups of features generated step by step. 

The first module generates statistical features ('first-order' features) based on the previous defined 'factor', 'key' and 'session' features, e.g. the number of unique values or the count of instances for a 'session' feature when grouped by the 'factor' feature. 
The first module can generate a small but effective group of new features, which show good performance in our experiments. 
We also call those generated features as \textit{baseline features}. 
Then, a feature selector is applied to the generated features to select the essential features to pass them to the following modules.

The second module generates 'second-order' combined features, which tries to combine the \textit{baseline features} with the other features. The operations vary in dealing with numerical or categorical features. 
For numerical features, we bin continuous data into intervals and calculate statistics on the data grouped by the 'key' features. 
For categorical features, various encoding methods, such as count/frequency encoding, are applied to the original columns to generate new features.
The experiments suggest that the features generated by the second module show different impacts on different datasets. 

The third module generates 'higher-order' features considering the temporal features. 
The temporal features are divided into several buckets, where new features are generated by operating on the previously generated features according to the time buckets. 
For example, the number of unique categories in a categorical feature is counted according to a specific period (seconds, minutes, hours, days, Etc.).
For the features generated by this module, we take similar approaches to select the essential features while choosing different numbers of features for different datasets according to the scale of the datasets. This is for saving memory usage while preserving favorable model performance.

The last module reviews all the categorical and multi-categorical features generated by previous steps and encodes them into learnable numerical values. 
A simple strategy for multi-categorical features is replacing the raw feature by taking the mean codes of the elements, e.g. replacing the set (0, 1, 2) by 1. 
For the categorical features, we record their label distribution over training samples.
For example, for a binary classification problem, the categorical value is replaced by the ratio of the corresponding positive and negative labels. 
The feature values for the testing samples are directly mapped from the ones for training samples.  
 
More details about the generated features can be found in our published GitHub repository~\footnote{\url{https://github.com/DeepBlueAI/AutoSmart}}. 

% The last module consists of target encoding and categorical mean encoding with supervised learning. The targe encoding is applied to categorical features, while categorical mean encoding is applied to multi-categorical features. The mean encoding values is simply the mean value of all the categorical code of a multi-categorical feature for each instance. The reason we execute this operation at the end is that it will generate too many features and run out of memory if it is executed in previous modules. 
% The last part consists of CTR and mean coding with supervised learning. The reason why we execute this operation at the end is that it will generate too many features and run out of memory.

\subsubsection{Feature Selection}
Feature engineering, when automated, can produce too many features causing extra time and memory computing cost, where feature selection is needed. Traditional ways include wrapper methods, e.g. recursive feature elimination (RFE)~\cite{guyon2002gene}, which greedily extracts the best-performed feature subset. They perform feature selection on all generated features, and each feature will be evaluated by a predictive model -- which causes external computation cost. 
We implement a multi-state feature pruning strategy to reduce the computation cost caused by feature generation and feature selection. 
As can be seen from Figure~\ref{fig:feature_engineering}, feature engineering is conducted recursively through multiple modules.  New features are generated from the main table at the beginning of each module, followed by a feature selection based on a down-sampled sub-dataset. The selected features are used to update the main table.  
Specifically, the information gain obtained from LightGBM is used to evaluate the feature importance in each feature selection part to remove irrelevant or partially irrelevant features.
Besides, we keep the lower-order features that can generate important higher-order features, so that the reconstructed main table contains informative and multi-perspective information.
This strategy prevents the feature generating modules from constructing useless high-order features in the early stages, thereby significantly reduce the generation time and memory usage. Furthermore, since we only do evaluations after each module, the time complexity of model training is reduced from $O(\#Features)$ to $O(\#Modules)$ compared with the RFE method.

\subsection{Automated Model Tuning}
We use LightGBM~\cite{ke2017lightgbm} as the prediction model for its lower memory usage, faster training speed, and higher efficiency. 
Two major hyper-parameters of the LightGBM model are the number of boosting modules and the learning rate for boosting modules. 
Most other teams used Bayesian Optimization~\cite{zhou2020kdd} for hyper-parameter tuning. However, this type of method needs training multiple times over the full samples to get the performance distribution of the hyper-parameters, which is time inefficient, especially when dealing with a large-scale dataset. 
Differently, we implement a wrapper-like approach with prior knowledge to reduce the search space.
For the number of boosting modules, we first implement two short trials with a small boost round number (we choose 15) and its double number. Then we could accurately calculate the real boost time by the time difference between these two trials, and further estimate the model preparation time and the max boost round according to the predefined time budget.
For the learning rate of the gradient boosting modules, we use sampled data to get the best learning rate through grid search and implement a customized exponential decay scheduler to achieve better results.
In summary, with the help of the sampled data or small boosting rounds, we successfully obtain necessary prior knowledge in a prior stage without training on the whole model. This strategy helps us achieve accurate results accompany the excellent time and resource control.

Besides, instead of using a single prediction model, we implement data-wise and feature-wise sampling methods to formulate an ensemble of prediction models to bring more randomness to the model. After testing the model performance by using bagging, blending, or stacking, we choose bagging as the ensemble strategy for its better performance and flexibility.
A more accurate and robust final prediction is achieved by simply averaging the predictions from these ensemble models.

%%%%%%%%%%%%%%%%%%%%%%%%%%%%%%%%%%%%%%%%%%%%%%%%%%%%%%%%%%%%%%%%%%%%%%%%%%%%%%%%%%%%%%%%%
\section{Experiments}
In the KDD CUP 2019 AutoML challenge, the target is developing AutoML solutions to binary classification problems for temporal relational data. 
The provided datasets are in the form of multiple related tables, with timestamped instances in real-world businesses. 
Five public datasets (without labels in the testing part) are provided for building the AutoML solutions. 
The solution will be further evaluated on five other unseen datasets without human intervention. The results of these five unseen datasets determine the final ranking. 
Each dataset is split into two subsets, namely the training set and the testing set. Both sets have: one main table file that stores the basic information; multiple related tables that store the auxiliary information; an info dictionary storing the feature types, the table relations, and a time budget (to run the whole framework); the training set has an additional label file that stores the labels associated with the main table. A basic description of the five public datasets is shown in Table~\ref{tab:dataset_description}. Besides, the allocated computational resources are a 4 Cores CPU with 16 GB memory for each submission.  
Thus, we need to provide the solution to the challenge by 1) effectively mining the useful information from the temporal relational data, 2) solving the class imbalance problem, and 3) minimizing the computation cost, all in an automatic way.
% for giving the best solution to the challenge, we need to: provide a good prediction result for mining the useful information from the temporal relational data, solve the class imbalance problem, and minimize the computation cost, all in an automatic way. 

\begin{table*}[]
    \centering
    \begin{tabular}{ccccccc}
    \toprule
        Dataset & \#Training Samples & \#Testing Samples & \#Features & Label Distribution & \#Related Tables & Time Budget (s) \\ \midrule
        SET1 & 808,953 & 779,656 & 37 & 7.11\% (Label 1) & 3 & 600\\
        SET2 & 1,111,133 & 1,000,160 & 23 & 2.25\% (Label 1) & 4 & 2400 \\
        SET3 & 1,003,286 & 981,609 & 13 & 0.06\% (Label 1) & 2 & 600 \\
        SET4 & 1,888,366 & 1,119,778 & 5 & 10.50\% (Label 0) & 3 & 1200 \\
        SET5 & 226,091 & 34,867 & 8 & 1.66\% (Label 1) & 3 & 300\\
    \bottomrule
    \end{tabular}
    \caption{A basic description of the five public datasets, where the label distribution is over training samples, the labels for testing samples are unavailable. }
    \label{tab:dataset_description}
\end{table*}

\begin{table*}[htbp]
  \caption{Ranking Score of KDD Cup 2019 AutoML Track at Feedback Phase on Five Public Datasets}
  \centering
  \setlength{\tabcolsep}{7mm}{
  \begin{tabular}{lrrrrrr}
    \toprule
    Team name & SET1 & SET2 & SET3 & SET4 & SET5 & AVG \\
    \midrule
    \textbf{AutoSmart (Ours)} & \bf1.0000 & \bf1.0000 & \bf1.0000 & 0.9871 & \bf1.0000 & \bf0.9974 \\
    Deep\_Wisdom & 0.8678 & 0.9167 & 0.9991 & \bf1.0000 & 0.9949 & 0.9557 \\
    MingLive & 0.6906 & 0.9296 & 0.4579 & 0.9487 & 0.2332 & 0.6520 \\
    xinjie & 0.6222 & 0.9641 & 0.3972 & 0.9471 & 0.2320 & 0.6325 \\
    yoshikawa & 0.6061 & 0.9357 & 0.4128 & 0.8438 & 0.2199 & 0.6037 \\
    bkdd & 0.6511 & 0.9485 & 0.3102 & 0.8608 & 0.1540 & 0.5849 \\
    ts302\_team & 0.4951 & 0.8240 & 0.2383 & 0.9275 & 0.2346 & 0.5439 \\
    PASA\_NJU & 0.4812 & 0.8483 & 0.3604 & 0.2370 & 0.1556 & 0.4165 \\
    TheRealRoman & 0.5889 & 0.9708 & 0.1617 & 0.8897 & 0.1205 & 0.5463 \\
    IIIS\_FLY & 0.5187 & 0.9283 & 0.2833 & 0.8365 & 0.1407 & 0.5415 \\
    \bottomrule
  \end{tabular}}
  \label{tab:results_1}
\end{table*}

\begin{table*}[htbp]
  \caption{Ranking Score of KDD Cup 2019 AutoML Track at Blind Test Phase on Five Private Datasets}
  \centering
  \setlength{\tabcolsep}{7mm}{
  \begin{tabular}{lrrrrrr}
    \toprule
    Team name & SET6 & SET7 & SET8 & SET9 & SET10 & AVG \\
    \midrule
    \textbf{AutoSmart (Ours)} & \bf1.0000 & \bf1.0000 & \bf1.0000 & 0.9287 & 0.6255 & \bf0.9108 \\
    xuechengxi & 0.3826 & 0.7548 & 0.2926 & \bf1.0000 & 0.6087 & 0.6077 \\
    admin & 0.4882 & 0.8441 & 0.3048 & 0.3461 & 0.9743 & 0.5915 \\
    ts302\_team & 0.5937 & 0.9943 & 0.6044 & 0.7501 & -0.1936 & 0.5498 \\
    yoshikawa & 0.7076 & 0.9101 & 0.4850 & 0.9440 & -0.3190 & 	0.5455 \\
    Apha & 0.2401 & 0.3921 & 0.2702 & 0.5921 & \bf1.0000 & 0.4989 \\
    teews & 0.4404 & 0.8615 & 0.4267 & 0.3204 & 0.4268 & 0.4952 \\
    PASA\_NJU & 0.5769 & 0.8880 & 0.6765 & 0.2422 & -0.0241 & 0.4719 \\
    \_CHAOS\_ & 0.4339 & 0.8867 & 0.5165 & 0.5509 & -0.2639 & 0.4248 \\
    IIIS\_FLY & 0.6236 & 0.9056 & 0.4236 & 0.9060 & -0.7895 & 0.4139 \\
    \bottomrule
  \end{tabular}}
  \label{tab:results_2}
\end{table*}

\begin{figure*}[t]
\centering
\includegraphics[width=0.95\textwidth]{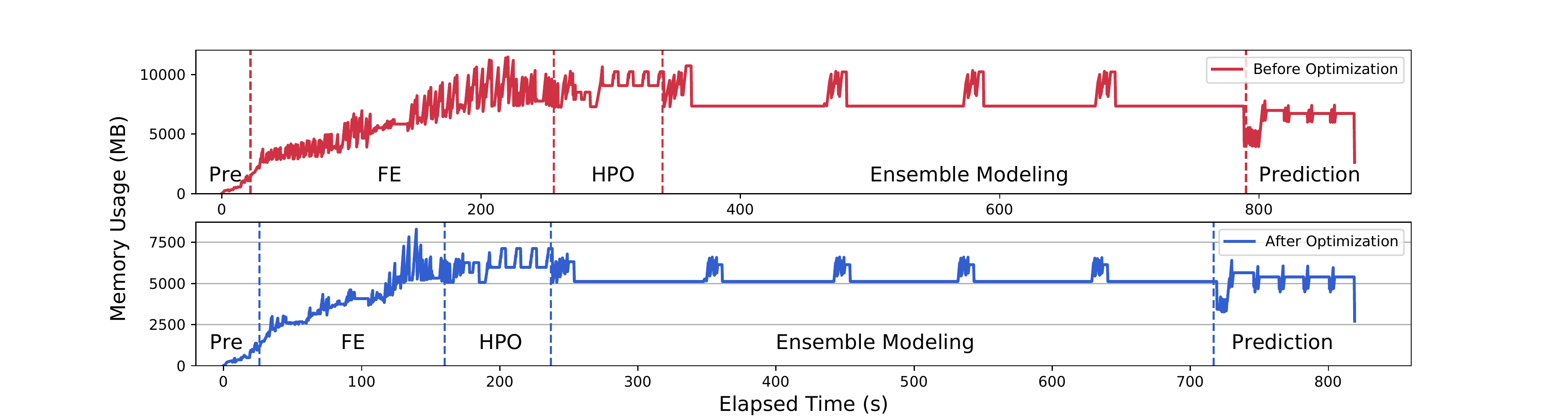}
\caption{Memory usage before and after optimization on SET 4 with a time budget of 1200 seconds. Time intervals for different modules are separated by dashed vertical lines. Pre, FE, and HPO stand for preprocessing, feature engineering and hyperparameter optimization.}
\label{fig:memory_usage}
\end{figure*}

\begin{figure}[t]
\centering
\includegraphics[width=0.95\linewidth]{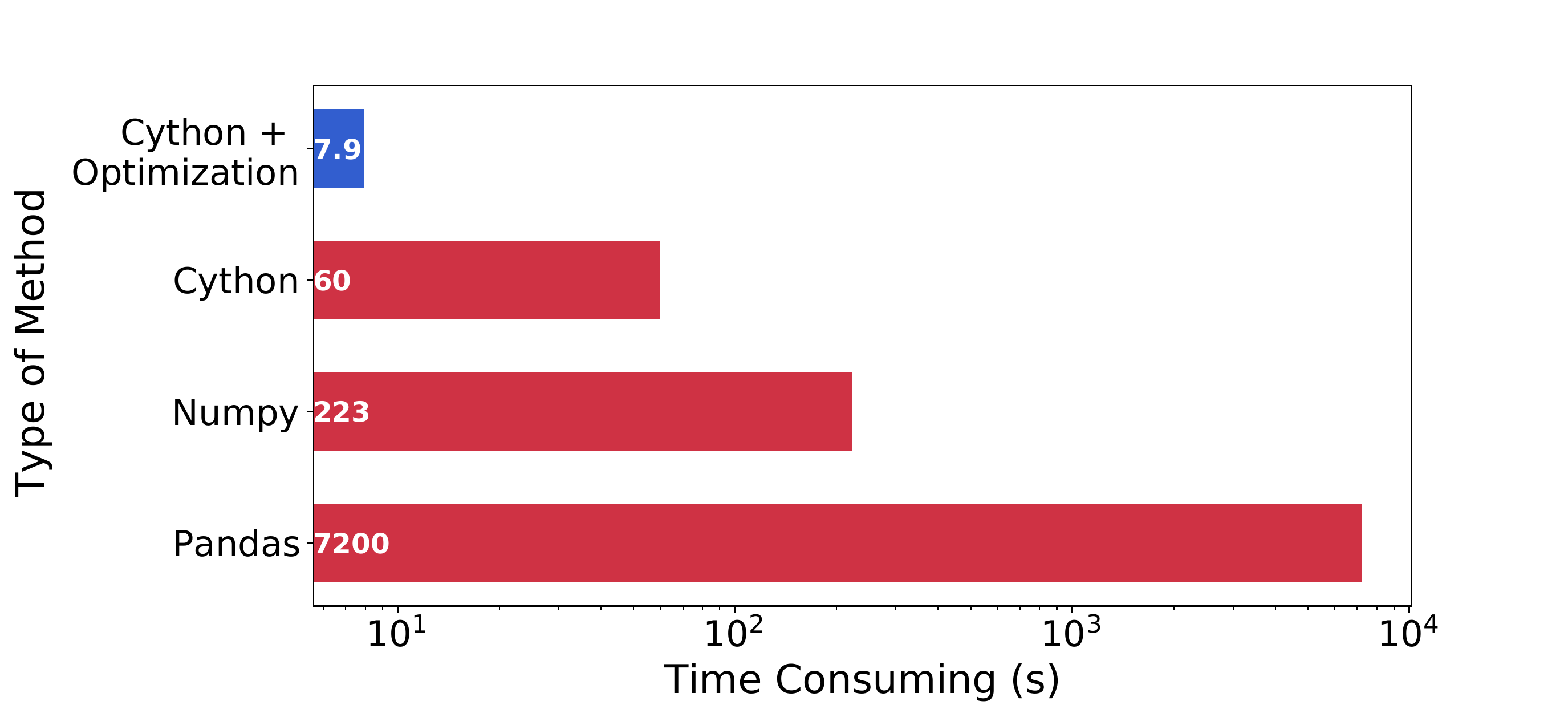}
\caption{Run time of the example multi-category feature generated by different methods, including Pandas, Numpy, Cython and Cython with optimization. We used Cython with optimization as the final selection in AutoSmart.}
\label{fig:time_consuming}
\end{figure}

\subsection{Experimental Performance}
% \begin{figure}
% \centering
% \includegraphics[width=0.85\linewidth]{autosmart/figures/auc_score.pdf}
% % \caption{Comparison of the performance of AutoSmart on private datasets at blind test phase with other top competitors}
% \caption{AUC of the top-3 winning solutions compared to baseline (by organizer) on five private datasets}
% \label{fig:auc_score}
% \end{figure}
The evaluation metric used in the KDD Cup 2019 challenge is calculated as:
\begin{equation}
score=\frac{auc-auc\_base}{auc\_max-auc\_base}
\label{eq1}
\end{equation}
where \textit{auc} is the resulting AUC (Area Under the Curve~\cite{fawcett2006introduction}) of the solution on the dataset, auc\_base is the AUC of the baseline method (provided by the sponsor) on the dataset, and auc\_max is the AUC of the best submitted solution on the dataset.
The average score on all five datasets (either public or private) will be used as the final score of a team, which is calculated as:
\begin{equation}
average\_score = \frac{1}{5}\sum_{i=1}^{5}score_i
\label{eq2}
\end{equation}
There are two phases, the feedback phase and the blind test phase, for evaluating the results. 
In the feedback phase, the participants can submit the solutions multiple times to get the test \textit{score} on five public datasets (Set 1 to 5). 
% the result \textit{scores} for the test datasets of the five public datasets (Set 1 to 5) are available on the platform, and the participants can submit the solutions multiple times. 
The last code submission of the feedback phase will be taken as the training code for the five unseen private datasets (Set 6 to 10) during the blind test phase. The performance for the five private datasets will not be revealed until the end of the contest.
Table~\ref{tab:results_1} and \ref{tab:results_2} present the evaluation \textit{scores} (calculated by Equation~\ref{eq1}) of the top 10 teams during the two phases, where we received 7 first places, 1 second place, and 2 third places for the ten datasets.
We can see that: \textbf{1)} the average score of our work exceeds the one from the second place by 4.14\% on the five public datasets, while this number increases to 30.31\% on the five private datasets; \textbf{2)} the results from most teams show unstable performance on different datasets; and \textbf{3)} some lead teams (e.g. Deep\_Wisdom and MingLive) who show good performance on the five public datasets do not succeed in obtaining similar performance on the five private datasets.
All those findings strongly demonstrate the effectiveness and the robustness of our proposed AutoML framework for temporal relational data, where we show excellent and stable performance over all datasets. 
% The organizer also provides a solution as the baseline solution, which also uses LightGBM for model training. The comparison between the top-3 winning solutions and the baseline methods is shown in Figure~\ref{fig:auc_score}. We can see that our proposed framework, AutoSmart, significantly outperforms the other solutions in nearly all scenarios.

\subsection{Resource Control}

The KDD challenge has a strict time and memory budget for each submission, which aims to provide fast AutoML solutions toward real-world applications. 
For building an efficient yet effective AutoML framework, we have taken time and memory control in nearly all processes in the framework.
The core ideas lie in under-sampling, code optimization, and risk assessment. 
We will introduce our general strategies in controlling time and memory usage and their performance in the following.

% For example, traditional feature engineering strategy will produce generated features (which can be a huge number of new features) at once, and select useful features among those features. 
% Differently, we divide feature engineering into multiple modules to sequentially generate and select features, which can significantly save the computation cost in model training. 
\subsubsection{Strategies in Time Control}
The time budget for training each dataset is normally hundreds of seconds, which means we need to control the time for generating each new feature within seconds for a table may containing millions of samples. 
For automatic data preprocessing and feature engineering, we use Cython~\footnote{\url{https://cython.org/}} to accelerate the process. 
We found that the processing of multi-categorical features is the bottleneck of controlling the feature engineering time to an acceptable rate during the experiments.
Figure~\ref{fig:time_consuming} shows the run time differences of several processing strategies for generating an example feature, which tries to identify the location of the value from one categorical feature in the set of another multi-categorical feature, e.g. [2137, (134, 2137, 576, 816)] -> 2. The processing time by directly using Pandas~\footnote{\url{https://pandas.pydata.org/}} API may take a few hours when a table contains millions of samples. 
The cost time can be decreased to minute level by using Numpy~\footnote{\url{https://numpy.org/}} or Cython, with the help of multi-threading, static typing, and more feasible data structure.
This time has been further optimized to seconds by using Cython with external arrays storing the value and length for the multi-categorical features. 

Model learning is another part that spends most of the training time. 
In our framework, we leverage the power of ensemble learning in constructing the models. 
% After testing the model performance by using bagging, blending, or stacking, we choose bagging as the ensemble strategy for building the predictors for its better performance and the flexibility. 
For saving the computation cost in ensemble learning, we firstly use a sample dataset to estimate the cost time in training the whole dataset by a single model, then calculate the remaining time to train other models. If the remaining time is not enough for building another single model, we early-stop the model training; if the remaining time is enough, we estimate the maximum number of models we can build to ensure the whole training process can successfully run within the time budget.
% time not exceed the time budget. 
% Correspondingly, our model can automatically fast adapt to the best situation given the time budget.
Figure~\ref{fig:memory_usage} gives an example of the model performance before and after memory control. We can see that the optimization in feature engineering reduces the processing time, where one more model is automatically added to the model ensembles to achieve better results.
% By taking advantage of this time saving, one more model is automatically added to the ensemble modeling to achieve a better result.

\subsubsection{Strategies in Memory Control}
To locate the memory usage peaks in the framework on different datasets, we implement a monitor to record the memory usage of a complete round of the system in advance. We found the memory peaks often appear at the merging of tables, the generation of certain features, and the start of the model training.
To avoid the memory explosion in the table merging and features generation, we adopt column-wise value assignment instead of using Pandas API, which may occupy twice the memory usage for the current data.
Similar to the measures in controlling the time usage, we estimate the maximum number of samples the system can take to meet the memory budget, so that to adjust the sample size and the feature size in building the prediction model.
% For saving the memory usage in model training, we take similar measures as in controlling the time usage, i.e. based on the resource taken by the sample dataset, we estimate the maximum number of samples the system can take for building an end-to-end framework while meeting the memory budget, and use the analytic results to adjust the sample size and the feature size in building the prediction model.
The selections of the subset in training the single models are shuffled in order to enhance the generalization ability of the ensembled model.
Another optimization is collecting the memory at the end of a variable life cycle, which can save the memory usage significantly.
Figure~\ref{fig:memory_usage} shows the memory usage of the whole process before and after our optimization strategies. Different parts of the framework are distinctly illustrated by the fluctuation of the monitored memory. The figure demonstrates that both peak memory and average memory are well controlled through the process, where the optimized peak memory consumption is capped by 8GB, a more than $30\%$ reduction compared to the original setting.

\subsection{Discussion}

\subsubsection{The class imbalance problem}
As shown in table~\ref{tab:dataset_description}, the label distribution of some of the datasets is highly imbalanced. For the case in SET3, the ratio of the label classes reaches as low as $0.06\%$. 
% To solve this problem, we adjust the ratio of label classes according to the size of the sampled data in training each model in ensemble models, so that we could keep more information from the original data as well as alleviate the class imbalance problem.
To solve this problem, we adjust the ratio of different labels according to the scale of the dataset to keep as much information as from the original data while alleviating the class imbalance problem.
Specifically, for the case that the ratio of minority and majority samples is smaller than 1:3, we apply under-sampling to the majority samples and increase the weights of minority samples.

\subsubsection{System Testing}
We test the system performance under some extreme conditions to check the robustness and the flexibility of the proposed AutoML framework.
First, the ability to handle rare conditions has been verified through a series of successful testings: missing values from different feature types; a column with all empty values; and a single table or many tables in a more complex structure.
Secondly, we evaluate the scalability of the system by testing the performance on larger datasets. For example, even we expand each data provided in this challenge by 2 times, 3 times, and 6 times, the system can still run smoothly. 
Lastly, to check if the system can self adapt the running time, we limit the time budget to 1/2, 1/3, or even 1/4 of the original one, where the system can still successfully produce accurate predictions. 
% All those testings show the efficiency, effectiveness, and the flexibility of the proposed model.

\section{Conclusions}
We propose an efficient and automatic machine learning framework, AutoSmart, for temporal relational data, which is the winning solution of the KDD Cup 2019 AutoML Track.
The framework includes automatic data processing, table merging, feature engineering, and model tuning, integrated with a time\&memory control unit. 
The experiments manifest that AutoSmart \textbf{1)} can effectively mining useful information and provide consistent outstanding performance on different temporal relational datasets; \textbf{2)} can efficiently self-tuning on the given datasets within the time and memory budget; \textbf{3)} is extendable to datasets with larger scales or some extreme cases (e.g. too many missing values). 
In a nutshell, the proposed framework can achieve the best and stable performance over different scenarios. Our code is publicly available, which is easy to reproduce and applied to industrial applications. 
% Future work could discuss how to time-efficiently address deep methods and utilize the semantic meanings of the categorical features.
Future research investigating how to apply deep methods to this temporal automatic framework in a timely and effective manner, as well as how to utilize the semantic meaning of categorical and text features, will be crucial.

% \begin{acks}
% The authors would like to thank HUANG ZHENDE, WU MENG QU for checking our English writing.
% \end{acks}

\bibliographystyle{ACM-Reference-Format}
\bibliography{reference}


\begin{thebibliography}{35}

%%% the Web address.

\ifx \showCODEN    \undefined \def \showCODEN     #1{\unskip}     \fi
\ifx \showDOI      \undefined \def \showDOI       #1{#1}\fi
\ifx \showISBNx    \undefined \def \showISBNx     #1{\unskip}     \fi
\ifx \showISBNxiii \undefined \def \showISBNxiii  #1{\unskip}     \fi
\ifx \showISSN     \undefined \def \showISSN      #1{\unskip}     \fi
\ifx \showLCCN     \undefined \def \showLCCN      #1{\unskip}     \fi
\ifx \shownote     \undefined \def \shownote      #1{#1}          \fi
\ifx \showarticletitle \undefined \def \showarticletitle #1{#1}   \fi
\ifx \showURL      \undefined \def \showURL       {\relax}        \fi
% invisible to TeX
\providecommand\bibfield[2]{#2}
\providecommand\bibinfo[2]{#2}
\providecommand\natexlab[1]{#1}
\providecommand\showeprint[2][]{arXiv:#2}

\bibitem[\protect\citeauthoryear{Baker, Gupta, Naik, and Raskar}{Baker
  et~al\mbox{.}}{2016}]%
        {baker2016designing}
\bibfield{author}{\bibinfo{person}{Bowen Baker}, \bibinfo{person}{Otkrist
  Gupta}, \bibinfo{person}{Nikhil Naik}, {and} \bibinfo{person}{Ramesh
  Raskar}.} \bibinfo{year}{2016}\natexlab{}.
\newblock \showarticletitle{Designing Neural Network Architectures using
  Reinforcement Learning}.
\newblock  (\bibinfo{year}{2016}).
\newblock


\bibitem[\protect\citeauthoryear{Bergstra, Bardenet, Bengio, and
  K{\'e}gl}{Bergstra et~al\mbox{.}}{2011}]%
        {bergstra2011algorithms}
\bibfield{author}{\bibinfo{person}{James Bergstra}, \bibinfo{person}{R{\'e}mi
  Bardenet}, \bibinfo{person}{Yoshua Bengio}, {and} \bibinfo{person}{Bal{\'a}zs
  K{\'e}gl}.} \bibinfo{year}{2011}\natexlab{}.
\newblock \showarticletitle{Algorithms for hyper-parameter optimization}. In
  \bibinfo{booktitle}{\emph{25th annual conference on neural information
  processing systems}}.
\newblock


\bibitem[\protect\citeauthoryear{Bisong}{Bisong}{2019}]%
        {bisong2019overview}
\bibfield{author}{\bibinfo{person}{Ekaba Bisong}.}
  \bibinfo{year}{2019}\natexlab{}.
\newblock \showarticletitle{An Overview of Google Cloud Platform Services}.
\newblock \bibinfo{journal}{\emph{Building Machine Learning and Deep Learning
  Models on Google Cloud Platform}} (\bibinfo{year}{2019}).
\newblock


\bibitem[\protect\citeauthoryear{Dash and Liu}{Dash and Liu}{1997}]%
        {dash1997feature}
\bibfield{author}{\bibinfo{person}{Manoranjan Dash} {and} \bibinfo{person}{Huan
  Liu}.} \bibinfo{year}{1997}\natexlab{}.
\newblock \showarticletitle{Feature Selection for Classification}.
\newblock  \bibinfo{volume}{1}, \bibinfo{number}{3} (\bibinfo{year}{1997}),
  \bibinfo{pages}{131--156}.
\newblock


\bibitem[\protect\citeauthoryear{Drori, Liu, Nian, Koorathota, Li, Moretti,
  Freire, and Udell}{Drori et~al\mbox{.}}{2019}]%
        {Drori2019AutoML}
\bibfield{author}{\bibinfo{person}{Iddo Drori}, \bibinfo{person}{Lu Liu},
  \bibinfo{person}{Yi Nian}, \bibinfo{person}{Sharath~C. Koorathota},
  \bibinfo{person}{Jie~S. Li}, \bibinfo{person}{Antonio~Khalil Moretti},
  \bibinfo{person}{Juliana Freire}, {and} \bibinfo{person}{Madeleine Udell}.}
  \bibinfo{year}{2019}\natexlab{}.
\newblock \showarticletitle{AutoML using Metadata Language Embeddings}.
\newblock  (\bibinfo{year}{2019}).
\newblock


\bibitem[\protect\citeauthoryear{Erickson, Mueller, Shirkov, Zhang, Larroy, Li,
  and Smola}{Erickson et~al\mbox{.}}{2020}]%
        {erickson2020autogluon}
\bibfield{author}{\bibinfo{person}{Nick Erickson}, \bibinfo{person}{Jonas
  Mueller}, \bibinfo{person}{Alexander Shirkov}, \bibinfo{person}{Hang Zhang},
  \bibinfo{person}{Pedro Larroy}, \bibinfo{person}{Mu Li}, {and}
  \bibinfo{person}{Alexander Smola}.} \bibinfo{year}{2020}\natexlab{}.
\newblock \showarticletitle{Autogluon-tabular: Robust and accurate automl for
  structured data}.
\newblock \bibinfo{journal}{\emph{arXiv preprint arXiv:2003.06505}}
  (\bibinfo{year}{2020}).
\newblock


\bibitem[\protect\citeauthoryear{Estevez-Velarde, Guti{\'e}rrez, Montoyo, and
  Almeida-Cruz}{Estevez-Velarde et~al\mbox{.}}{2019}]%
        {estevez2019automl}
\bibfield{author}{\bibinfo{person}{Suilan Estevez-Velarde},
  \bibinfo{person}{Yoan Guti{\'e}rrez}, \bibinfo{person}{Andr{\'e}s Montoyo},
  {and} \bibinfo{person}{Yudivi{\'a}n Almeida-Cruz}.}
  \bibinfo{year}{2019}\natexlab{}.
\newblock \showarticletitle{AutoML strategy based on grammatical evolution: A
  case study about knowledge discovery from text}. In
  \bibinfo{booktitle}{\emph{Proceedings of the 57th Annual Meeting of the
  Association for Computational Linguistics}}. \bibinfo{pages}{4356--4365}.
\newblock


\bibitem[\protect\citeauthoryear{Fawcett}{Fawcett}{2006}]%
        {fawcett2006introduction}
\bibfield{author}{\bibinfo{person}{Tom Fawcett}.}
  \bibinfo{year}{2006}\natexlab{}.
\newblock \showarticletitle{An introduction to ROC analysis}.
\newblock \bibinfo{journal}{\emph{Pattern recognition letters}}
  \bibinfo{volume}{27}, \bibinfo{number}{8} (\bibinfo{year}{2006}),
  \bibinfo{pages}{861--874}.
\newblock


\bibitem[\protect\citeauthoryear{Feurer, Klein, Eggensperger, Springenberg,
  Blum, and Hutter}{Feurer et~al\mbox{.}}{2019}]%
        {feurer2019auto}
\bibfield{author}{\bibinfo{person}{Matthias Feurer}, \bibinfo{person}{Aaron
  Klein}, \bibinfo{person}{Katharina Eggensperger},
  \bibinfo{person}{Jost~Tobias Springenberg}, \bibinfo{person}{Manuel Blum},
  {and} \bibinfo{person}{Frank Hutter}.} \bibinfo{year}{2019}\natexlab{}.
\newblock \showarticletitle{Auto-sklearn: efficient and robust automated
  machine learning}.
\newblock In \bibinfo{booktitle}{\emph{Automated Machine Learning}}.
\newblock


\bibitem[\protect\citeauthoryear{Graepel, Lauter, and Naehrig}{Graepel
  et~al\mbox{.}}{2012}]%
        {graepel2012ml}
\bibfield{author}{\bibinfo{person}{Thore Graepel}, \bibinfo{person}{Kristin
  Lauter}, {and} \bibinfo{person}{Michael Naehrig}.}
  \bibinfo{year}{2012}\natexlab{}.
\newblock \showarticletitle{ML confidential: Machine learning on encrypted
  data}. In \bibinfo{booktitle}{\emph{International Conference on Information
  Security and Cryptology}}. Springer, \bibinfo{pages}{1--21}.
\newblock


\bibitem[\protect\citeauthoryear{Guyon, Weston, Barnhill, and Vapnik}{Guyon
  et~al\mbox{.}}{2002}]%
        {guyon2002gene}
\bibfield{author}{\bibinfo{person}{Isabelle Guyon}, \bibinfo{person}{Jason
  Weston}, \bibinfo{person}{Stephen Barnhill}, {and} \bibinfo{person}{Vladimir
  Vapnik}.} \bibinfo{year}{2002}\natexlab{}.
\newblock \showarticletitle{Gene selection for cancer classification using
  support vector machines}.
\newblock \bibinfo{journal}{\emph{Machine learning}} \bibinfo{volume}{46},
  \bibinfo{number}{1} (\bibinfo{year}{2002}), \bibinfo{pages}{389--422}.
\newblock


\bibitem[\protect\citeauthoryear{Hassan, Smith, Xu, Zhai, Veeramachaneni,
  et~al\mbox{.}}{Hassan et~al\mbox{.}}{2020}]%
        {hassan2020level}
\bibfield{author}{\bibinfo{person}{Md Hassan}, \bibinfo{person}{Micah~J Smith},
  \bibinfo{person}{Lei Xu}, \bibinfo{person}{ChengXiang Zhai},
  \bibinfo{person}{Kalyan Veeramachaneni}, {et~al\mbox{.}}}
  \bibinfo{year}{2020}\natexlab{}.
\newblock \showarticletitle{A Level-wise Taxonomic Perspective on Automated
  Machine Learning to Date and Beyond: Challenges and Opportunities}.
\newblock \bibinfo{journal}{\emph{arXiv preprint arXiv:2010.10777}}
  (\bibinfo{year}{2020}).
\newblock


\bibitem[\protect\citeauthoryear{He, Zhao, and Chu}{He et~al\mbox{.}}{2019}]%
        {He2019AutoML}
\bibfield{author}{\bibinfo{person}{Xin He}, \bibinfo{person}{Kaiyong Zhao},
  {and} \bibinfo{person}{Xiaowen Chu}.} \bibinfo{year}{2019}\natexlab{}.
\newblock \showarticletitle{AutoML: A Survey of the State-of-the-Art}.
\newblock  (\bibinfo{year}{2019}).
\newblock


\bibitem[\protect\citeauthoryear{He, Lin, Liu, Wang, Li, and Han}{He
  et~al\mbox{.}}{2018}]%
        {he2018amc}
\bibfield{author}{\bibinfo{person}{Yihui He}, \bibinfo{person}{Ji Lin},
  \bibinfo{person}{Zhijian Liu}, \bibinfo{person}{Hanrui Wang},
  \bibinfo{person}{Li-Jia Li}, {and} \bibinfo{person}{Song Han}.}
  \bibinfo{year}{2018}\natexlab{}.
\newblock \showarticletitle{Amc: Automl for model compression and acceleration
  on mobile devices}. In \bibinfo{booktitle}{\emph{Proceedings of the European
  Conference on Computer Vision (ECCV)}}. \bibinfo{pages}{784--800}.
\newblock


\bibitem[\protect\citeauthoryear{Hutter, Hoos, and Leyton-Brown}{Hutter
  et~al\mbox{.}}{2014}]%
        {Hutter2014An}
\bibfield{author}{\bibinfo{person}{Frank Hutter}, \bibinfo{person}{Holger
  Hoos}, {and} \bibinfo{person}{Kevin Leyton-Brown}.}
  \bibinfo{year}{2014}\natexlab{}.
\newblock \showarticletitle{An efficient approach for assessing hyperparameter
  importance}. In \bibinfo{booktitle}{\emph{International Conference on Machine
  Learning}}.
\newblock


\bibitem[\protect\citeauthoryear{Kanter and Veeramachaneni}{Kanter and
  Veeramachaneni}{2015}]%
        {Kanter2015Deep}
\bibfield{author}{\bibinfo{person}{James~Max Kanter} {and}
  \bibinfo{person}{Kalyan Veeramachaneni}.} \bibinfo{year}{2015}\natexlab{}.
\newblock \showarticletitle{Deep feature synthesis: Towards automating data
  science endeavors}. In \bibinfo{booktitle}{\emph{IEEE International
  Conference on Data Science and Advanced Analytics}}.
\newblock


\bibitem[\protect\citeauthoryear{Katz, Shin, and Song}{Katz
  et~al\mbox{.}}{2016}]%
        {Katz2016ExploreKit}
\bibfield{author}{\bibinfo{person}{Gilad Katz}, \bibinfo{person}{Eui
  Chul~Richard Shin}, {and} \bibinfo{person}{Dawn Song}.}
  \bibinfo{year}{2016}\natexlab{}.
\newblock \showarticletitle{ExploreKit: Automatic Feature Generation and
  Selection}. In \bibinfo{booktitle}{\emph{2016 IEEE 16th International
  Conference on Data Mining (ICDM)}}.
\newblock


\bibitem[\protect\citeauthoryear{Ke, Meng, Finley, Wang, Chen, Ma, Ye, and
  Liu}{Ke et~al\mbox{.}}{2017}]%
        {ke2017lightgbm}
\bibfield{author}{\bibinfo{person}{Guolin Ke}, \bibinfo{person}{Qi Meng},
  \bibinfo{person}{Thomas Finley}, \bibinfo{person}{Taifeng Wang},
  \bibinfo{person}{Wei Chen}, \bibinfo{person}{Weidong Ma},
  \bibinfo{person}{Qiwei Ye}, {and} \bibinfo{person}{Tie-Yan Liu}.}
  \bibinfo{year}{2017}\natexlab{}.
\newblock \showarticletitle{Lightgbm: A highly efficient gradient boosting
  decision tree}.
\newblock \bibinfo{journal}{\emph{Advances in neural information processing
  systems}}  \bibinfo{volume}{30} (\bibinfo{year}{2017}),
  \bibinfo{pages}{3146--3154}.
\newblock


\bibitem[\protect\citeauthoryear{Khurana, Turaga, Samulowitz, and
  Parthasrathy}{Khurana et~al\mbox{.}}{2016}]%
        {Khurana2016Cognito}
\bibfield{author}{\bibinfo{person}{Udayan Khurana}, \bibinfo{person}{Deepak
  Turaga}, \bibinfo{person}{Horst Samulowitz}, {and}
  \bibinfo{person}{Srinivasan Parthasrathy}.} \bibinfo{year}{2016}\natexlab{}.
\newblock \showarticletitle{Cognito: Automated Feature Engineering for
  Supervised Learning}. In \bibinfo{booktitle}{\emph{IEEE International
  Conference on Data Mining Workshops}}.
\newblock


\bibitem[\protect\citeauthoryear{Klein}{Klein}{2017}]%
        {klein2017azure}
\bibfield{author}{\bibinfo{person}{Scott Klein}.}
  \bibinfo{year}{2017}\natexlab{}.
\newblock \showarticletitle{Azure machine learning}.
\newblock In \bibinfo{booktitle}{\emph{IoT Solutions in Microsoft's Azure IoT
  Suite}}. \bibinfo{publisher}{Springer}, \bibinfo{pages}{227--252}.
\newblock


\bibitem[\protect\citeauthoryear{K{\"o}ppen}{K{\"o}ppen}{2000}]%
        {koppen2000curse}
\bibfield{author}{\bibinfo{person}{Mario K{\"o}ppen}.}
  \bibinfo{year}{2000}\natexlab{}.
\newblock \showarticletitle{The curse of dimensionality}. In
  \bibinfo{booktitle}{\emph{5th Online World Conference on Soft Computing in
  Industrial Applications (WSC5)}}, Vol.~\bibinfo{volume}{1}.
  \bibinfo{pages}{4--8}.
\newblock


\bibitem[\protect\citeauthoryear{Lam, Thiebaut, Sinn, Chen, Mai, and Alkan}{Lam
  et~al\mbox{.}}{2017}]%
        {lam2017one}
\bibfield{author}{\bibinfo{person}{Hoang~Thanh Lam},
  \bibinfo{person}{Johannmichael Thiebaut}, \bibinfo{person}{Mathieu Sinn},
  \bibinfo{person}{Bei Chen}, \bibinfo{person}{Tiep Mai}, {and}
  \bibinfo{person}{Oznur Alkan}.} \bibinfo{year}{2017}\natexlab{}.
\newblock \showarticletitle{One button machine for automating feature
  engineering in relational databases.}
\newblock \bibinfo{journal}{\emph{arXiv: Databases}} (\bibinfo{year}{2017}).
\newblock


\bibitem[\protect\citeauthoryear{LeDell and Poirier}{LeDell and
  Poirier}{2020}]%
        {ledell2020h2o}
\bibfield{author}{\bibinfo{person}{Erin LeDell} {and} \bibinfo{person}{S
  Poirier}.} \bibinfo{year}{2020}\natexlab{}.
\newblock \showarticletitle{H2o automl: Scalable automatic machine learning}.
  In \bibinfo{booktitle}{\emph{7th ICML workshop on automated machine
  learning}}.
\newblock


\bibitem[\protect\citeauthoryear{Margaritis}{Margaritis}{2009}]%
        {Margaritis2009Toward}
\bibfield{author}{\bibinfo{person}{Dimitris Margaritis}.}
  \bibinfo{year}{2009}\natexlab{}.
\newblock \showarticletitle{Toward Provably Correct Feature Selection in
  Arbitrary Domains}. In \bibinfo{booktitle}{\emph{Advances in Neural
  Information Processing Systems 22: Conference on Neural Information
  Processing Systems A Meeting Held December}}.
\newblock


\bibitem[\protect\citeauthoryear{Olson and Moore}{Olson and Moore}{2016}]%
        {olson2016tpot}
\bibfield{author}{\bibinfo{person}{Randal~S Olson} {and}
  \bibinfo{person}{Jason~H Moore}.} \bibinfo{year}{2016}\natexlab{}.
\newblock \showarticletitle{TPOT: A tree-based pipeline optimization tool for
  automating machine learning}. In \bibinfo{booktitle}{\emph{Workshop on
  automatic machine learning}}. PMLR, \bibinfo{pages}{66--74}.
\newblock


\bibitem[\protect\citeauthoryear{Sagi and Rokach}{Sagi and Rokach}{2018}]%
        {sagi2018ensemble}
\bibfield{author}{\bibinfo{person}{Omer Sagi} {and} \bibinfo{person}{Lior
  Rokach}.} \bibinfo{year}{2018}\natexlab{}.
\newblock \showarticletitle{Ensemble learning: A survey}.
\newblock \bibinfo{journal}{\emph{Wiley Interdisciplinary Reviews: Data Mining
  and Knowledge Discovery}} \bibinfo{volume}{8}, \bibinfo{number}{4}
  (\bibinfo{year}{2018}), \bibinfo{pages}{e1249}.
\newblock


\bibitem[\protect\citeauthoryear{Shahriari, Swersky, Wang, Adams, and
  De~Freitas}{Shahriari et~al\mbox{.}}{2015}]%
        {shahriari2015taking}
\bibfield{author}{\bibinfo{person}{Bobak Shahriari}, \bibinfo{person}{Kevin
  Swersky}, \bibinfo{person}{Ziyu Wang}, \bibinfo{person}{Ryan~P Adams}, {and}
  \bibinfo{person}{Nando De~Freitas}.} \bibinfo{year}{2015}\natexlab{}.
\newblock \showarticletitle{Taking the human out of the loop: A review of
  Bayesian optimization}.
\newblock \bibinfo{journal}{\emph{Proc. IEEE}} \bibinfo{volume}{104},
  \bibinfo{number}{1} (\bibinfo{year}{2015}), \bibinfo{pages}{148--175}.
\newblock


\bibitem[\protect\citeauthoryear{Shaul, MarkovitchDan, and Rosenstein}{Shaul
  et~al\mbox{.}}{2002}]%
        {Shaul2002Feature}
\bibfield{author}{\bibinfo{person}{Shaul}, \bibinfo{person}{MarkovitchDan},
  {and} \bibinfo{person}{Rosenstein}.} \bibinfo{year}{2002}\natexlab{}.
\newblock \showarticletitle{Feature Generation Using General Constructor
  Functions}.
\newblock \bibinfo{journal}{\emph{Machine Learning}} (\bibinfo{year}{2002}).
\newblock


\bibitem[\protect\citeauthoryear{Snoek, Larochelle, and Adams}{Snoek
  et~al\mbox{.}}{2012}]%
        {snoek2012practical}
\bibfield{author}{\bibinfo{person}{Jasper Snoek}, \bibinfo{person}{Hugo
  Larochelle}, {and} \bibinfo{person}{Ryan~P Adams}.}
  \bibinfo{year}{2012}\natexlab{}.
\newblock \showarticletitle{Practical bayesian optimization of machine learning
  algorithms}.
\newblock \bibinfo{journal}{\emph{arXiv preprint arXiv:1206.2944}}
  (\bibinfo{year}{2012}).
\newblock


\bibitem[\protect\citeauthoryear{Wang, Ming, Jin, Shen, Liu, Smith,
  Veeramachaneni, and Qu}{Wang et~al\mbox{.}}{2019}]%
        {Wang2019ATMSeer}
\bibfield{author}{\bibinfo{person}{Qianwen Wang}, \bibinfo{person}{Yao Ming},
  \bibinfo{person}{Zhihua Jin}, \bibinfo{person}{Qiaomu Shen},
  \bibinfo{person}{Dongyu Liu}, \bibinfo{person}{Micah~J Smith},
  \bibinfo{person}{Kalyan Veeramachaneni}, {and} \bibinfo{person}{Huamin Qu}.}
  \bibinfo{year}{2019}\natexlab{}.
\newblock \showarticletitle{ATMSeer: Increasing Transparency and
  Controllability in Automated Machine Learning}.
\newblock  (\bibinfo{year}{2019}).
\newblock


\bibitem[\protect\citeauthoryear{Wong, Houlsby, Lu, and Gesmundo}{Wong
  et~al\mbox{.}}{2018}]%
        {wong2018transfer}
\bibfield{author}{\bibinfo{person}{Catherine Wong}, \bibinfo{person}{Neil
  Houlsby}, \bibinfo{person}{Yifeng Lu}, {and} \bibinfo{person}{Andrea
  Gesmundo}.} \bibinfo{year}{2018}\natexlab{}.
\newblock \showarticletitle{Transfer learning with neural automl}.
\newblock \bibinfo{journal}{\emph{Neural Information Processing Systems}}
  (\bibinfo{year}{2018}).
\newblock


\bibitem[\protect\citeauthoryear{Yu and Zhu}{Yu and Zhu}{2020}]%
        {yu2020hyper}
\bibfield{author}{\bibinfo{person}{Tong Yu} {and} \bibinfo{person}{Hong Zhu}.}
  \bibinfo{year}{2020}\natexlab{}.
\newblock \showarticletitle{Hyper-parameter optimization: A review of
  algorithms and applications}.
\newblock \bibinfo{journal}{\emph{arXiv preprint arXiv:2003.05689}}
  (\bibinfo{year}{2020}).
\newblock


\bibitem[\protect\citeauthoryear{Zhou, Roy, and Skrypnyk}{Zhou
  et~al\mbox{.}}{2020}]%
        {zhou2020kdd}
\bibfield{author}{\bibinfo{person}{Wenjun Zhou}, \bibinfo{person}{Taposh~Dutta
  Roy}, {and} \bibinfo{person}{Iryna Skrypnyk}.}
  \bibinfo{year}{2020}\natexlab{}.
\newblock \showarticletitle{The KDD Cup 2019 Report}.
\newblock \bibinfo{journal}{\emph{ACM SIGKDD Explorations Newsletter}}
  \bibinfo{volume}{22}, \bibinfo{number}{1} (\bibinfo{year}{2020}),
  \bibinfo{pages}{8--17}.
\newblock


\bibitem[\protect\citeauthoryear{Zller and Huber}{Zller and Huber}{2020}]%
        {Marc2020Benchmark}
\bibfield{author}{\bibinfo{person}{Marc-Andre Zller} {and}
  \bibinfo{person}{Marco~F. Huber}.} \bibinfo{year}{2020}\natexlab{}.
\newblock \showarticletitle{Benchmark and Survey of Automated Machine Learning
  Frameworks}.
\newblock  (\bibinfo{year}{2020}).
\newblock


\bibitem[\protect\citeauthoryear{Z{\"o}ller and Huber}{Z{\"o}ller and
  Huber}{2019}]%
        {zoller2019benchmark}
\bibfield{author}{\bibinfo{person}{Marc-Andr{\'e} Z{\"o}ller} {and}
  \bibinfo{person}{Marco~F Huber}.} \bibinfo{year}{2019}\natexlab{}.
\newblock \showarticletitle{Benchmark and survey of automated machine learning
  frameworks}.
\newblock \bibinfo{journal}{\emph{arXiv preprint arXiv:1904.12054}}
  (\bibinfo{year}{2019}).
\newblock


\end{thebibliography}

\end{document}